\documentclass[10pt,twocolumn,letterpaper]{article}

\usepackage{cvpr}
\usepackage{times}
\usepackage{epsfig}
\usepackage{graphicx}
\usepackage{amsmath}
\usepackage{amssymb}
\usepackage{graphicx}  \graphicspath{{figures/}}
\usepackage{algorithm}
\usepackage{algpseudocode}
\usepackage[normalem]{ulem}
\usepackage{makecell}
\usepackage{rotating}
\usepackage{booktabs}
\usepackage{gensymb}
\usepackage{capt-of}
\usepackage{enumitem}
\usepackage[dvipsnames,svgnames,x11names]{xcolor}

\usepackage[pagebackref=true,breaklinks=true,letterpaper=true,colorlinks,bookmarks=false]{hyperref}
\usepackage[toc,page]{appendix}

\cvprfinalcopy 

\def\cvprPaperID{704} 

\newcommand{\setmode}[1]{\def\mode{#1}}
\setmode{draft} 
\long\def\IGNORE#1{} \long\def\COMMENT#1{}

\ifthenelse{\equal{\mode}{draft}} { 
    \def\authornote#1#2#3{{\textcolor{#2}{\textsl{\small#1:[*#3*]}}}}
}
{
    \def\authornote#1#2#3{}
    \typeout{************* MODE: Final}
}

\newcommand{\mysubsubsection}[1]{\vspace{0.1cm} \noindent {\bf #1}:}

\newcolumntype{L}[1]{>{\raggedright\let\newline\\\arraybackslash\hspace{0pt}}m{#1}}
\newcolumntype{C}[1]{>{\centering\let\newline\\\arraybackslash\hspace{0pt}}m{#1}}
\newcolumntype{R}[1]{>{\raggedleft\let\newline\\\arraybackslash\hspace{0pt}}m{#1}}

\newcommand{\fig}[1]{Fig.~\ref{#1}}
\newcommand{\tbl}[1]{Table~\ref{#1}}

\newcommand{\ignore}[1]{}

\makeatletter
\DeclareRobustCommand\onedot{\futurelet\@let@token\@onedot}
\def\@onedot{\ifx\@let@token.\else.\null\fi\xspace}

\def\etal{\emph{et al}\onedot}

\makeatother

\definecolor{MyDarkBlue}{rgb}{0,0.08,1}
\definecolor{MyDarkGreen}{rgb}{0.02,0.6,0.02}
\definecolor{MyDarkRed}{rgb}{0.8,0.02,0.02}
\definecolor{MyDarkOrange}{rgb}{0.40,0.2,0.02}
\definecolor{MyPurple}{RGB}{111,0,255}
\definecolor{MyRed}{rgb}{1.0,0.0,0.0}
\definecolor{MyGold}{rgb}{0.75,0.6,0.12}
\definecolor{MyDarkgray}{rgb}{0.66, 0.66, 0.66}

\ifcvprfinal\fi

\title{PlaneRCNN: 3D Plane Detection and Reconstruction from a Single Image}

\author{Chen Liu$^{1,2}$\footnotemark[1]
\qquad Kihwan Kim$^1$ \qquad Jinwei Gu$^{1,3}$\footnotemark[1] \qquad Yasutaka Furukawa$^4$ \qquad Jan Kautz$^1$\\
$^1$NVIDIA \qquad $^2$Washington University in St. Louis\\
$^3$SenseTime \qquad $^4$Simon Fraser University \\
}

\begin{document}

\twocolumn[{
\maketitle
\centerline{
\includegraphics[width=\textwidth,trim={4pt 4pt 4pt 4pt}]{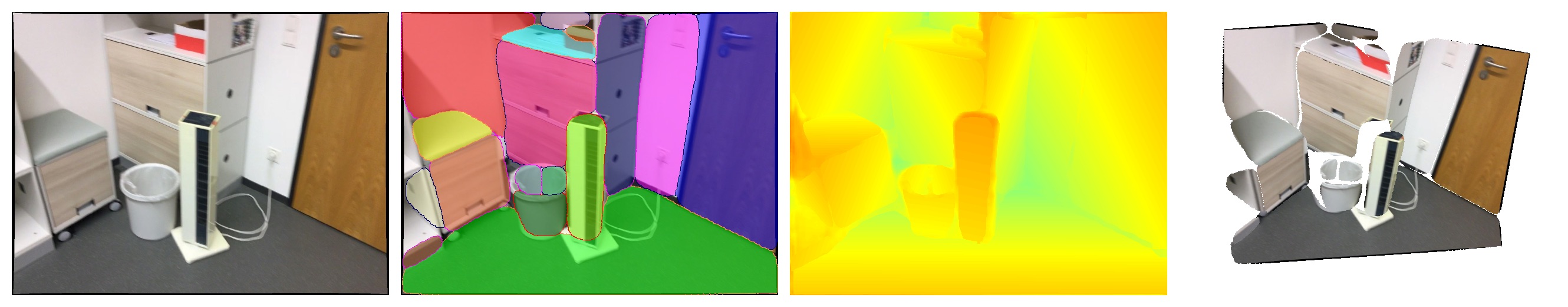}}
\captionof{figure}
{
This paper proposes a deep neural architecture, PlaneRCNN, that detects planar regions and reconstructs a piecewise planar depthmap from a single RGB image.
From left to right, an input image, segmented planar regions, estimated depthmap, and reconstructed planar surfaces.
}
\label{fig:teaser}
\vspace{2em}
}]



\begin{abstract}
This paper proposes a deep neural architecture, PlaneRCNN, that detects and reconstructs piecewise planar surfaces from a single RGB image. PlaneRCNN employs a variant of Mask R-CNN to detect planes with their plane parameters and segmentation masks. PlaneRCNN then jointly refines
all the segmentation masks with a novel loss enforcing the consistency with a nearby view during training. The paper also presents a new benchmark with more fine-grained plane segmentations in the ground-truth, in which, PlaneRCNN outperforms existing state-of-the-art methods with significant margins in the plane detection, segmentation, and reconstruction metrics. PlaneRCNN makes an important step towards robust plane extraction, which would have an immediate impact on a wide range of applications including Robotics, Augmented Reality, and Virtual Reality.  
\end{abstract}

\section{Introduction}
\label{sec::intro}
\footnotetext[1]{The authors contributed to this work when they were at NVIDIA.}
Planar regions in 3D scenes offer important geometric cues in a variety of 3D perception tasks such as scene understanding~\cite{Tsai11}, scene reconstruction~\cite{Chauve10}, and robot navigation~\cite{Kaess15, Zhou06}. 
Accordingly, piecewise planar scene reconstruction has been a focus of computer vision research for many years, for example, 
plausible recovery of planar structures from a single image~\cite{Hoiem05}, volumetric piecewise planar reconstruction from point clouds~\cite{Chauve10},
and Manhattan depthmap reconstruction from multiple images~\cite{Furukawa09}. 

A difficult yet fundamental task is the inference of a piecewise planar structure from a single RGB image, posing two key challenges.
First, 3D plane reconstruction from a single image is an ill-posed problem, requiring rich scene priors. Second, planar structures abundant in man-made environments often lack textures, requiring global image understanding as opposed to local texture analysis.
%
Recently, PlaneNet~\cite{liu2018planenet} and PlaneRecover~\cite{yang2018recovering} made a breakthrough 
by introducing the use of Convolutional Neural Networks (CNNs) and formulating the problem as a plane segmentation task.
%
While generating promising results, they suffer from three major limitations: 1) Missing small surfaces; 2) Requiring the maximum number of planes in a single image a priori; and 3) Poor generalization across domains (e.g., trained for indoors images and tested outdoors).


This paper proposes a novel deep neural architecture, PlaneRCNN, that addresses these issues and more effectively
infers piecewise planar structure from a single RGB image (\fig{fig:teaser}). PlaneRCNN consists of three components.
%

The first component is a plane detection network built upon Mask R-CNN~\cite{he2017mask}. Besides an instance mask for each planar region, we also estimate the plane normal and per-pixel depth values. With known camera intrinsics, we can further reconstruct the 3D planes from the detected planar regions.
%
%
This detection framework is more flexible and can handle an arbitrary number of planar regions in an image.
To the best of our knowledge, this paper is the first to introduce a detection network, common in object recognition, to the 
depthmap reconstruction task.
%
The second component is a segmentation refinement network that jointly optimizes extracted segmentation masks to more coherently explain a scene as a whole. The refinement network is designed to handle an arbitrary number of regions via a simple yet effective neural module. 
The third component, the warping-loss module, enforces the consistency of reconstructions with another view observing the same scene \emph{during training} and improves the plane parameter and depthmap accuracy in the detection network via end-to-end training.
%

The paper also presents a new benchmark for the piecewise planar depthmap reconstruction task. We collected 100,000 images from ScanNet~\cite{dai2017scannet} and generated the corresponding 
ground-truth by utilizing the associated 3D scans.
The new benchmark offers 14.7 plane instances per image on the average, in contrast to roughly 6 instances per image in the existing benchmark~\cite{liu2018planenet}.
%

The performance is evaluated via plane detection, segmentation, and reconstruction metrics, in which PlaneRCNN outperforms the current state-of-the-art with significant margins. Especially, PlaneRCNN is able to detect small planar surfaces and generalize well to new scene types.

The contributions of the paper are two-fold:\\
\noindent \textbf{Technical Contribution}:
The paper proposes a novel neural architecture PlaneRCNN, where 1) a detection network extracts an arbitrary number of planar regions; 2) a refinement network jointly improves all the segmentation masks; and 3) a warping loss improves plane-parameter and depthmap accuracy via end-to-end training.

\noindent \textbf{System Contribution}: 
The paper provides a new benchmark for the piecewise planar depthmap reconstruction task with much more fine-grained annotations than before, in which
PlaneRCNN makes significant improvements over the current state-of-the-art.

\section{Related Work}

For 3D plane detection and reconstruction, most traditional approaches~\cite{furukawa2009manhattan,gallup2010piecewise,silberman2012indoor,sinha2009piecewise,zebedin2008fusion} require multiple views or depth information as input. They generate plane proposals by fitting planes to 3D points, then assign a proposal to each pixel via a global inference.
Deng~\etal~\cite{deng2017unsupervised} proposed a learning-based approach to recover planar regions, while still requiring depth information as input.

Recently, PlaneNet~\cite{liu2018planenet} revisited the piecewise planar depthmap reconstruction problem with an end-to-end learning framework from a single indoor RGB image.
PlaneRecover~\cite{yang2018recovering} 
later proposed an un-supervised learning approach for outdoor scenes. Both PlaneNet and PlaneRecover formulated the task as a pixel-wise segmentation problem with a fixed number of planar regions (i.e., $10$ in PlaneNet and $5$ in PlaneRecover), which severely limits the expressiveness of their reconstructions and generalization capabilities to different scene types.
%
We address these limitations by utilizing a detection network, commonly used for object recognition.


Detection-based framework has been successfully applied to many 3D understanding tasks for objects, for example, predicting object shapes in the form of bounding boxes~\cite{chen2016monocular,fidler20123d,mousavian20173d}, wire-frames~\cite{li2016deep,wu2016single,zia2013detailed}, or template-based shape compositions~\cite{chabot2017deep,kundu18,mottaghi2015coarse,xiang2015data}.
%
%
However, the coarse representation employed in these methods lack the ability to accurately model complex and cluttered indoor scenes.

In addition to the detection, joint refinement of segmentation masks is also a key to many applications that require precise plane parameters or boundaries.
In recent 
semantic segmentation techniques,
fully connected conditional random field (CRF) is proven to be effective for localizing segmentation boundaries~\cite{chen2014semantic,krahenbuhl2011efficient}.
CRFasRNN~\cite{zheng2015conditional} further makes it differentiable for end-to-end training. 
CRF only utilizes low-level information, and
global context is further exploited via
RNNs~\cite{byeon2015scene,liang2016semantic,shuai2016dag}, 
more general graphical models~\cite{liu2015semantic,lin2016efficient}, or novel neural architectural designs~\cite{zhang2017global,zhao2017pyramid,yu2015multi}.
These segmentation refinement techniques are NOT instance-aware, merely inferring a semantic label at each pixel and cannot distinguish multiple instances belonging to the same semantic category.

Instance-aware joint segmentation refinement poses more challenges.
Traditional methods~\cite{sun2014relating,tighe2013finding,tighe2014scene,tu2005image,yao2012describing} model the scene as a graph and use graphical model inference techniques to jointly optimize all instance masks. With a sequence of heuristics, these methods are often not robust.
To this end, we will propose a segmentation refinement network that jointly optimizes an arbitrary number of segmentation masks on top of a detection network.

\section{Approach}
\label{sec:approach}

\setcounter{figure}{1} 
\begin{figure*}[t]
	\centering
    \includegraphics[width=.99\linewidth]{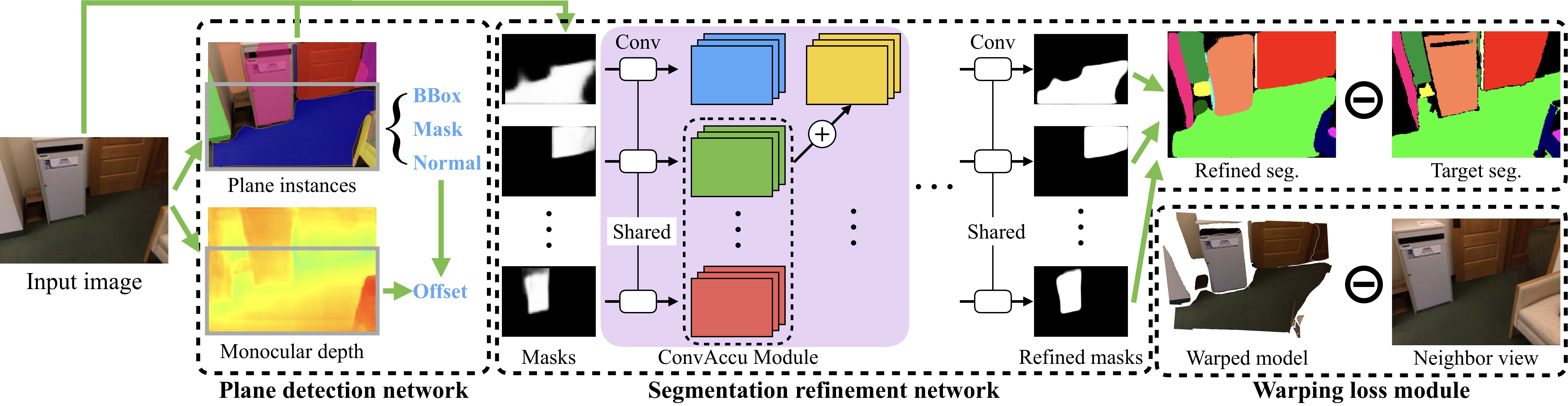}
    \caption{Our framework consists of three building blocks: 1) a plane detection network based on Mask R-CNN~\cite{he2017mask}, 2) a segmentation refinement network that jointly optimizes extracted segmentation masks,
    and 3) a warping loss module that enforces the consistency of reconstructions with a nearby view during training.
    }
	\label{fig:pipeline}
\end{figure*}


PlaneRCNN consists of three main components (See \fig{fig:pipeline}):
a plane detection network, a segmentation refinement network, and a warping loss module. Built upon Mask R-CNN~\cite{he2017mask}, the plane proposal network (Sec.~\ref{sec:planeproposal}) detects planar regions given a single RGB image and predicts 3D plane parameters together with a segmentation mask for each planar region. The refinement network (Sec.~\ref{sec:planerefine}) takes all detected planar regions and jointly optimizes their masks. 
The warping loss module (Sec.~\ref{sec:two-view-consistency}) enforces the consistency of reconstructed planes with another view observing the same scene to further improve the accuracy of plane parameters and depthmap during training.


\subsection{Plane Detection Network}
\label{sec:planeproposal}

Mask R-CNN was originally designed for semantic segmentation, where
images contain instances of varying categories (e.g., person, car, train, bicycle and more).
Our problem has only two categories "planar" or "non-planar", defined in a geometric sense.
Nonetheless,
Mask R-CNN works surprisingly well in detecting planes in our experiments.
%
%
It also enables us to handle 
an arbitrary number of planes, where existing approaches need the maximum number of planes in an image a priori (i.e., 10 for PlaneNet~\cite{liu2018planenet} and 5 for PlaneRecover~\cite{yang2018recovering}).
%

We treat each planar region as an object instance and let Mask R-CNN detect such instances and estimate their segmentation masks. The remaining task is to infer 3D plane parameters, which consists of the normal and the offset information. 
%
While CNNs have been successful for depthmap~\cite{liu2016learning} and surface normal~\cite{wang2015designing} estimation, direct regression of plane offset turns out to be a challenge (even with the use of CoordConv~\cite{liu2018intriguing}). 
Instead of direct regression, we solve it in three steps: (1) predict a normal per planar instance, (2) estimate a depthmap for an entire image, and (3) use a simple algebraic formula (Eq.~\ref{eq:offset}) to calculate the plane offset (which is differentiable for end-to-end training).
%
%
We now explain how we modify Mask-RCNN to perform these three steps. 

\IGNORE{
\footnote{However, the ROI pooling scheme in Mask R-CNN discards the location information of the pooled feature map. 
One could add the image coordinates as an additional feature layer, for example as shown in CoordConv~\cite{liu2018intriguing}, but we found this method does not solve the problem.}
}


\mysubsubsection{Plane normal estimation}
Directly attaching a parameter regression module after the ROI pooling produces reasonable results, but we borrow the idea of 2D anchor boxes for bounding box regression~\cite{he2017mask} to further improve accuracy.
%
More precisely, we consider \emph{anchor normals} and estimate a plane normal in the local camera coordinate frame by 1) picking an anchor normal, 2) regressing the residual 3D vector, and 3) normalizing the sum to a unit-length vector.

\begin{figure}[h]
	\centering
    \includegraphics[width=.99\linewidth]{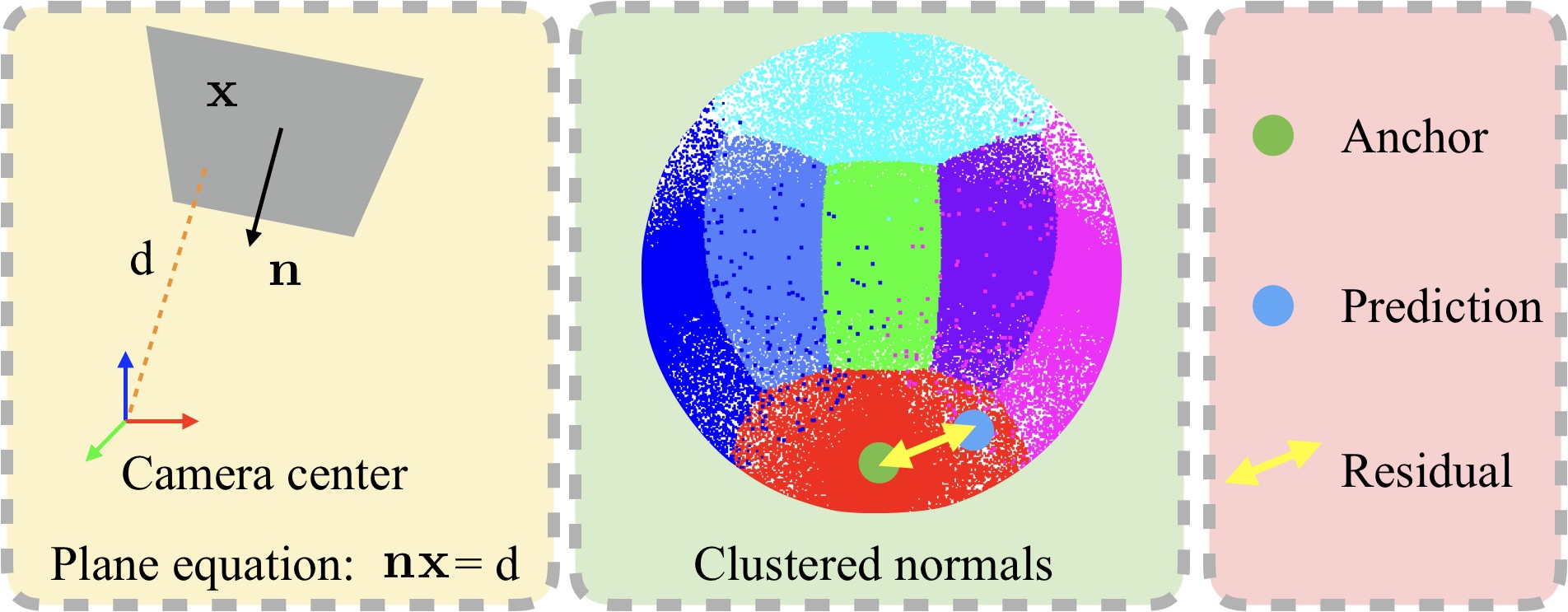}
\caption{We estimate a plane normal by first picking one of the 7 anchor normals and then regressing the residual 3D vector. Anchor normals are defined by running the K-means clustering algorithm on the ground-truth plane normal vectors.
}
%
	\label{fig:normals}
\end{figure}

Anchor normals are defined by running the K-means clustering algorithm on the plane normals in $10,000$ randomly sampled training images. 
We use $k=7$ and the cluster centers become anchor normals, which are
up-facing, down-facing, and horizontal vectors roughly separated by $45^\circ$ in our experiments (See~\fig{fig:normals}).

We replace the object category prediction in the original Mask R-CNN with the anchor ID prediction, and append one separate fully-connected layer
to regress the 3D residual vector for each anchor normal (i.e., $21=3\times7$ output values). To generate supervision for each ground-truth plane normal, we find the closest anchor normal and compute the residual vector. We use the cross-entropy loss for the anchor normal selection, and the smooth L1 loss for the residual vector regression as in the bounding box regression of Mask R-CNN.
%

\mysubsubsection{Depthmap estimation}
While local image analysis per region suffices for surface normal prediction, global image analysis is crucial for depthmap inference.
We add a decoder after the feature pyramid network (FPN)~\cite{lin2017feature} in Mask R-CNN to estimate
the depthmap $D$ for an entire image. 
%
For the depthmap decoder, we use a block of $3 \times 3$ convolution with stride $1$ and $4 \times 4$ deconvolution with stride $2$ at each level. Lastly, bilinear upsampling is used to generate a depthmap in the same resolution as the input image ($640 \times 640$).

\IGNORE{To be specific, FPN features maps have resolutions 
$[10 \times 10, 20 \times 20, 40 \times 40, 80 \times 80, 160 \times 160]$. The depth inference decoder 
uses $3 \times 3$ convolution with stride $1$ and$4 \times 4$ deconvolution with stride $2$ to process the first feature map from the encoder, whose output is concatenated to the second feature map.  
We repeat the same to process all the feature maps before applying bilinear upsampling to get a pixel-wise depthmap of the same resolution as the input image ($640 \times 640$).
}

\mysubsubsection{Plane offset estimation}
Given a plane normal $\mathbf{n}$, it is straightforward to estimate the plane offset $d$:
\begin{equation}
d=\frac{\sum_i{m_i(\mathbf{n^{\intercal}}(z_iK^{-1}\mathbf{x_i}))}}{\sum_i{m_i}}
\label{eq:offset}
\end{equation}
where $K$ is the $3\times 3$ camera intrinsic matrix, $\mathbf{x_i}$ is the $i_{\mbox{th}}$ pixel coordinate in a homogeneous representation, $z_i$ is its predicted depth value, and $m_i$ is an indicator variable, which becomes 1 if the pixel belongs to the plane. The summation is over all the pixels in the image. Note that we do not have a loss on the plane offset parameter, which did not make differences in the results. However, the plane offset influences the warping loss module below.

\subsection{Segmentation Refinement Network}
\label{sec:planerefine}
The plane detection network predicts segmentation masks independently. 
The segmentation refinement network jointly optimizes all the masks, where
the major challenge lies in the varying number of detected planes.
One solution is to assume the maximum number of planes in an image, concatenate all the masks, and pad zero in the missing entries.
However, this does not scale to a large number of planes, and is prone to missing small planes.
%

Instead, we propose a simple yet effective module, ConvAccu, based on the idea of non-local module~\cite{wang2018non}.
ConvAccu processes each plane segmentation mask represented in the entire image window with a convolution layer. We then calculate and concatenate the mean feature volumes over all the other planes at the same level before passing to the next level (See~\fig{fig:pipeline}). This resembles the non-local module and can effectively aggregate information from all the masks. We built an U-Net~\cite{ronneberger2015u} architecture using ConvAccu modules with details illustrated in Appendix~\ref{appendix:refinement}.
%

Refined plane masks are concatenated at the end and compared against ground-truth with a cross-entropy loss.
Note that besides the plane mask, the refinement network also takes the original image, the union of all the other plane masks, the reconstructed depthmap (for planar and non-planar regions), and a 3D coordinate map for the specific plane as input. 
%
The target segmentation mask is generated on the fly during training by assigning a ground-truth mask with the largest overlap.
Planes without any assigned ground-truth masks do not receive supervision.
%

\subsection{Warping Loss Module}
\label{sec:two-view-consistency}
The warping loss module enforces the consistency of reconstructed 3D planes with a nearby view during training.
Specifically, our training samples come from RGB-D videos in ScanNet~\cite{dai2017scannet}, and the nearby view is defined to be the one 20 frames ahead from the current.
%
The module first builds a depthmap for each frame by 1) computing depth values from the plane equations for planar regions and 2) using pixel-wise depth values predicted inside the plane detection network for the remaining pixels. Depthmaps are converted to 3D coordinate maps in the local camera coordinate frames (i.e., a 3D coordinate instead of a depth value per pixel) by using the camera intrinsic information.

The warping loss is then computed as follows. Let $M_c$ and $M_n$ denote the 3D coordinate maps of the current and the nearby frames, respectively.
For every 3D point $\mathbf{p}_n(\in M_n)$ in the nearby view, we use the camera pose information to project to the current frame, and use a bilinear interpolation to read the 3D coordinate $\mathbf{p}_c$ from $M_c$. We then transform $\mathbf{p}_c$ to the coordinate frame of the nearby view based on the camera pose and compute the 3D distance between the transformed coordinate $\mathbf{p}_c^t$ and $\mathbf{p}_n$. L2 norm of all such 3D distances divided by the number of pixels is the loss.
We ignore pixels that project outside the current image frame during bilinear interpolation.
%

The projection, un-projection, and coordinate frame transformation are all simple algebraic operations, whose gradients can be passed for training. Note that the warping loss module and the nearby view is utilized only during training to boost geometric reconstruction accuracy, and the system runs on a single image at test time.

\section{Benchmark construction}
Following steps described in PlaneNet~\cite{liu2018planenet}, we build a new benchmark from RGB-D videos in ScanNet~\cite{dai2017scannet}.
We add the following three modifications to recover more fine-grained planar regions, yielding $14.7$ plane instances per image on the average, which is more than double the PlaneNet dataset containing $6.0$ plane instances per image.

%
%

\vspace{0.1cm}
\noindent $\bullet$ First, we keep more small planar regions by reducing the plane area threshold from $1\%$ of the image size to $0.16\%$ (i.e., 500 pixels) and not dropping small planes when the total number is larger than $10$.

\vspace{0.1cm}
\noindent $\bullet$ Second, PlaneNet merges co-planar planes into a single region as they share the same plane label. The merging of two co-planar planes from different objects causes loss of semantics. We skip the merging process and keep all instance segmentation masks.

\vspace{0.1cm}
\noindent $\bullet$ Third, the camera pose quality in ScanNet degrades in facing 3D tracking failures, which causes misalignment between image and the projected ground-truth planes.
Since we use camera poses and aligned 3D models to generate ground-truth planes, we detect such failures by the discrepancy between our ground-truth 3D planes and the raw depthmap from a sensor. 
More precisely, we do not use images if the average depth discrepancy over planar regions is larger than 0.1m. This simple strategy removes approximately $10\%$ of the images.




\section{Experimental results}
We have implemented our network in PyTorch. We use pre-trained Mask R-CNN~\cite{he2017mask} and initialize the segmentation refinement network with the existing model~\cite{he2015delving}. We train the network end-to-end on an NVIDIA TitanX GPU for 10 epochs with 100,000 randomly sampled images from training scenes in ScanNet. We use the same scale factor for all losses. For the detection network, we scale the image to $640 \times 480$ and pad zero values to get a $640 \times640$ input image. For the refinement network, we scale the image to $256 \times 192$ and align the detected instance masks with the image based on the predicted bounding boxes.


\begin{figure}[t]
	\centering
    \includegraphics[width=.99\linewidth]{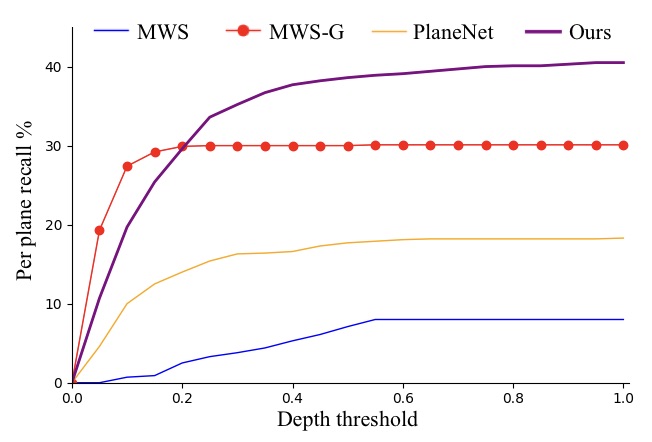}
\caption{Plane-wise accuracy against baselines. PlaneRCNN outperforms all the competing methods except when the depth threshold is very small and MWS-G can fit 3D planes extremely accurately by utilizing the ground-truth depth values.
%
}
	\label{fig:curves}
\end{figure}

\subsection{Qualitative evaluations}
\label{sect:qualitative}
\begin{figure*}[htb]
\begin{minipage}[t]{0.48\textwidth}
    \includegraphics[width=\linewidth]{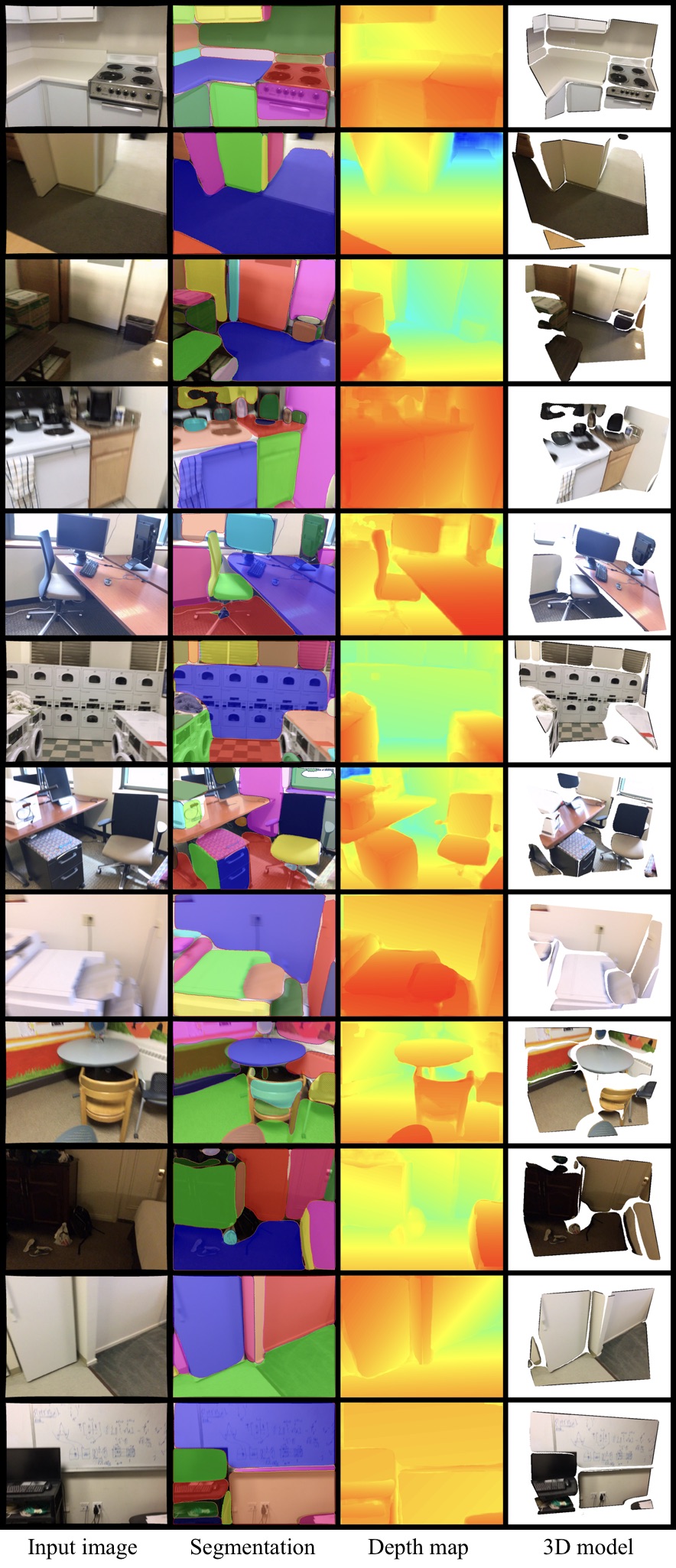}
\caption{Piecewise planar reconstruction results by PlaneRCNN. From left to right: input image, plane segmentation, depthmap reconstruction, and 3D rendering of our depthmap (rendered from a new view with -0.4m and 0.3m translation along x-axis and z-axis respectively and $10^{\degree}$ rotation along both x-axis and z-axis).}
    \label{fig:results}
\end{minipage}%
%
\hfill%
\begin{minipage}[t]{0.48\textwidth}
    \includegraphics[width=\linewidth]{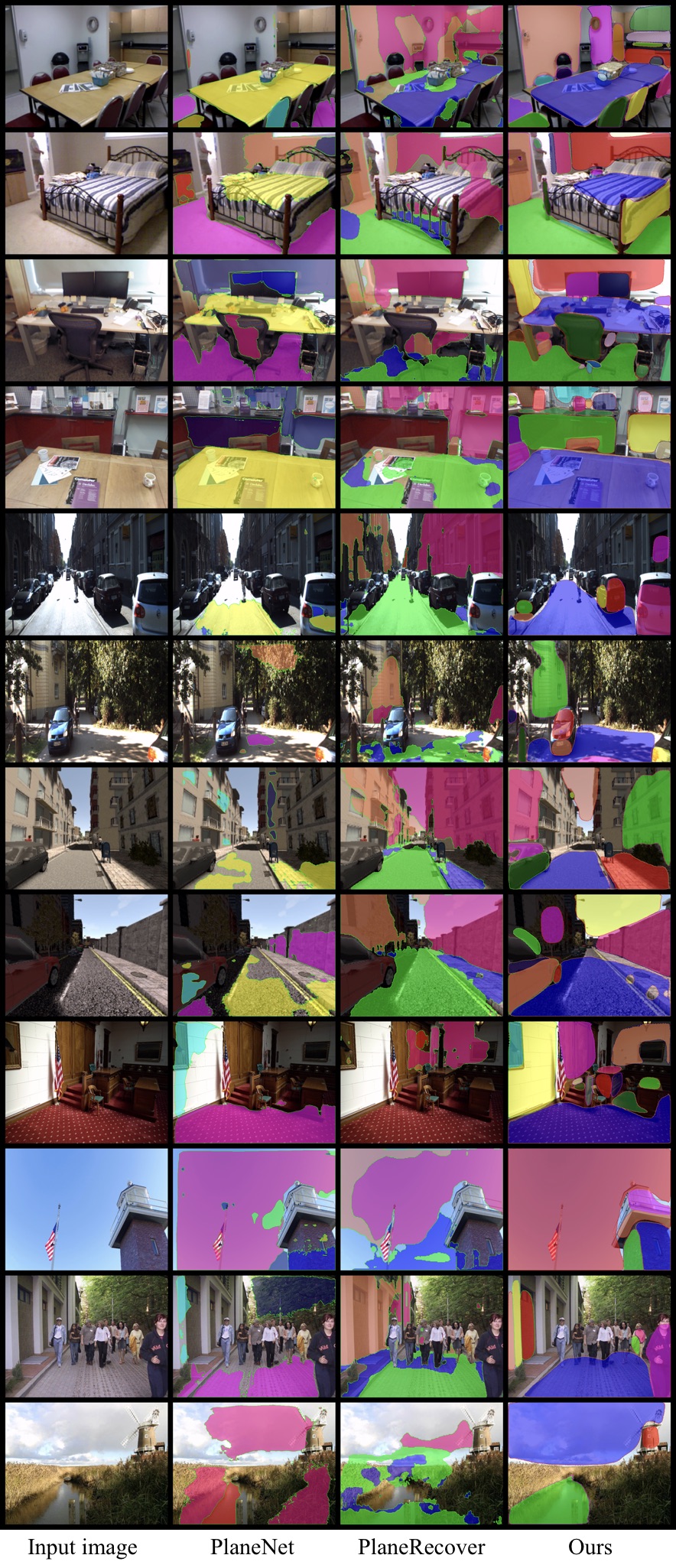}
\caption{Plane segmentation results on unseen datasets without fine-tuning. From left to right: input image, PlaneNet~\cite{liu2018planenet} results, PlaneRecover~\cite{yang2018recovering} results, and ours. From top to the bottom, we show two examples from each dataset in the order of NYUv2~\cite{silberman2012indoor}, 7-scenes~\cite{shotton2013scene}, KITTI~\cite{geiger2013vision}, SYNTHIA~\cite{ros2016synthia}, Tank and Temple~\cite{knapitsch2017tanks}, and PhotoPopup~\cite{hoiem2005automatic}.
}
	\label{fig:cross_dataset}
\end{minipage}
\end{figure*}

Fig.~\ref{fig:results} demonstrates our reconstructions results for ScanNet testing scenes. PlaneRCNN is able to recover planar surfaces even for small objects.
We include more examples in Appendix~\ref{appendix:results}.

Fig.~\ref{fig:cross_dataset} compares PlaneRCNN against two competing methods, PlaneNet~\cite{liu2018planenet} and PlaneRecover~\cite{yang2018recovering}, on a variety of scene types from unseen datasets (except the SYNTHIA dataset is used for training by PlaneRecover).
%
Note that PlaneRCNN and PlaneNet are trained on the ScanNet which contains indoor scenes, while PlaneRecover is trained on the SYNTHIA dataset (i.e., the 7th and 8th rows in the figure) which consist of synthetic outdoor scenes. 
The figure shows that PlaneRCNN is able to reconstruct most planes in varying scene types from unseen datasets regardless of their sizes, shapes, and textures.
%
In particular,
our results on the KITTI dataset are surprisingly better than PlaneRecover for planes close to the camera. In indoor scenes, our results are consistently better than both PlaneNet and PlaneRecover.
We include more examples in Appendix~\ref{appendix:results}.

\subsection{Plane reconstruction accuracy}
\label{sect:reconstruction}
Following PlaneNet~\cite{liu2018planenet}, we evaluate plane detection accuracy by measuring the plane recall with a fixed Intersection over Union (IOU) threshold $0.5$ and a varying depth error threshold (from $0$ to $1m$ with an increment of $0.05m$).
The accuracy is measured inside the overlapping regions between the ground-truth and inferred planes. Besides PlaneNet, we compare against Manhattan World Stereo (MWS)~\cite{furukawa2009manhattan}, which is the most competitive traditional MRF-based approach as demonstrated in prior evaluations~\cite{liu2018planenet}. 
MWS requires a 3D point cloud as an input, and we either use the point cloud from the ground-truth 3D planes (MWS-G) or the point cloud inferred by our depthmap estimation module in the plane detection network (MWS). 
%
PlaneRecover~\cite{yang2018recovering} was originally trained with the assumption of at most 5 planes in an image. We find it difficult to train PlaneRecover successfully for cluttered indoor scenes by simply increasing the threshold.
We believe that PlaneNet, which is explicitly trained on ScanNet, serves as a stronger competitor for the evaluation.

\begin{figure*}
	\centering
	\newlength\exlen
    \setlength\exlen{\linewidth}
    \includegraphics[width=\linewidth]{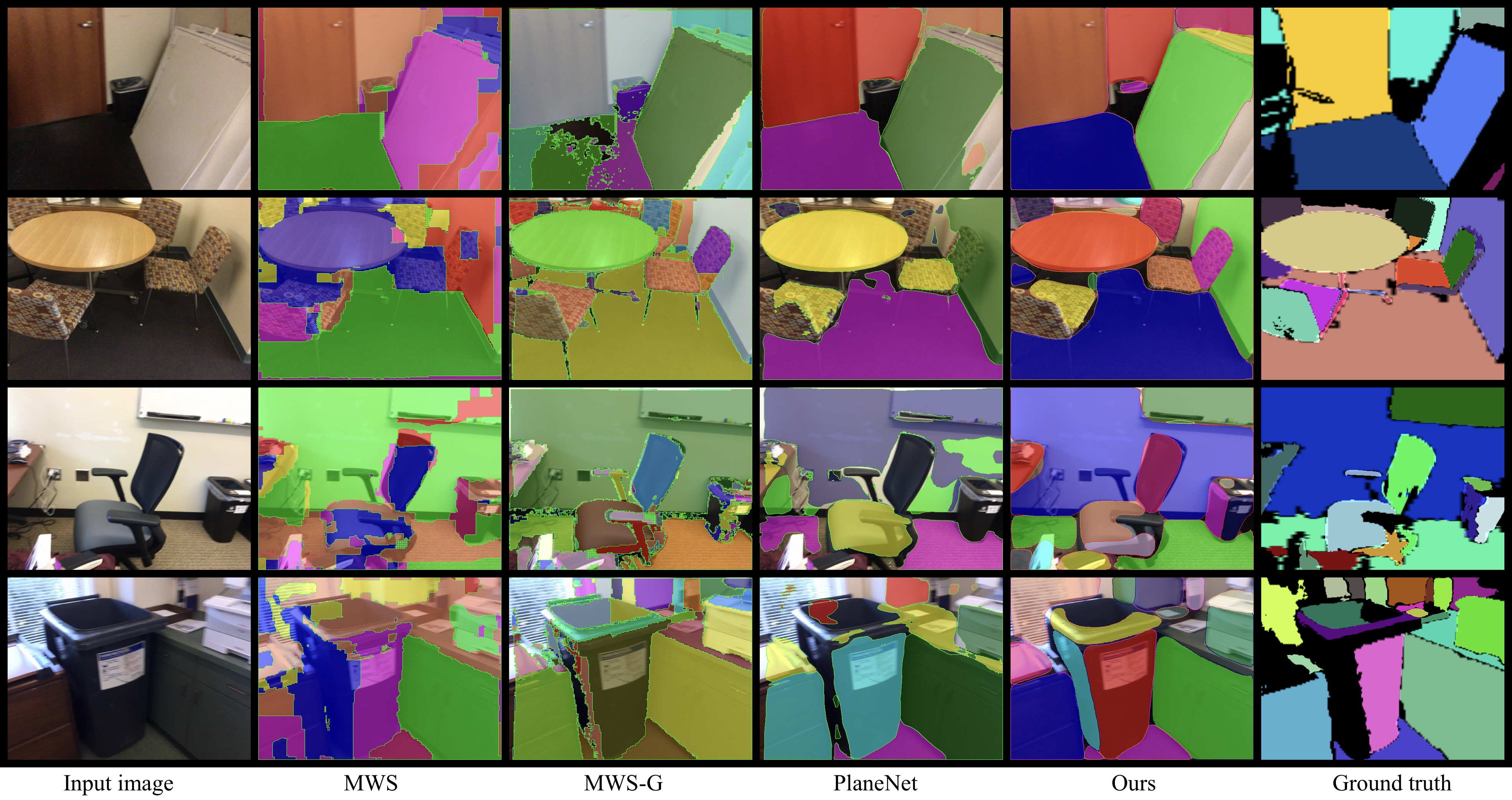}
\vspace{2pt}
\caption{Plane segmentation comparisons.
From left to right: input image, MWS~\cite{furukawa2009manhattan} with inferred depths, MWS~\cite{furukawa2009manhattan} with ground-truth depths, PlaneNet~\cite{liu2018planenet}, Ours, and ground-truth.}
	\label{fig:comparison}
	\vspace{3pt}
\end{figure*}

As demonstrated in \fig{fig:curves}, PlaneRCNN significantly outperforms all other methods,
except when the depth threshold is small and MWS-G can fit planes extremely accurately with the ground-truth depth values.
Nonetheless, even with ground-truth depth information, MWS-G fails in extracting planar regions robustly, leading to lower recalls in general.
Our results are superior also qualitatively as shown 
in~\fig{fig:comparison}.


\subsection{Geometric accuracy}
\label{sect:geometry}
We propose a new metric in evaluating the quality of piecewise planar surface reconstruction by mixing the inferred depthmaps and the ground-truth plane segmentations. More precisely, we first 
generate a depthmap from our reconstruction by following the process in the warping loss evaluation (Sec.~\ref{sec:two-view-consistency}).
%
%
Next, for every ground-truth planar segment, we convert depth values in the reconstructed depthmap to 3D points, fit a 3D plane by SVD, and normalize the plane coefficients to make the normal component into a unit vector.
Finally, 
we compute the mean and the area-weighted mean of the parameter differences to serve as the evaluation metrics. Besides the plane parameter metrics, we also consider depthmap metrics commonly used in the literature~\cite{eigen2015predicting}. We evaluate over the NYU dataset~\cite{silberman2012indoor} for a fair comparison.
Table~\ref{tbl:comparison} shows that, with more flexible detection network, PlaneRCNN generalizes much better
without fine-tuning. PlaneRCNN also outperforms PlaneNet~\cite{liu2018planenet} in every metric after fine-tuning using the ground-truth depths from the NYU dataset.


\begin{table}[h]
\caption{Geometric accuracy comparison over the NYUv2 dataset.
}
\label{tbl:comparison}
  \centering
  \begin{tabular}{l|c|c}
    \toprule
  Method & PlaneNet~\cite{liu2018planenet} & Ours \\
  \midrule
  \multicolumn{3}{c}{w/o fine-tuning} \\
  \midrule
  Rel & 0.220 & \textbf{0.164} \\
  $log_{10}$ & 0.114 & \textbf{0.077} \\
  $RMSE$ & 0.858 & \textbf{0.644} \\
  Param. & 0.939 & \textbf{0.776} \\
  Param. (weighted) & 0.771 & \textbf{0.641} \\
 \midrule
  \multicolumn{3}{c}{w/ fine-tuning} \\
  \midrule
  Rel & 0.129 & \textbf{0.124} \\
  $log_{10}$ & 0.079 & \textbf{0.073} \\
  $RMSE$ & 0.397 & \textbf{0.395} \\
  Param. & 0.713 & \textbf{0.642} \\
  Param. (weighted) & 0.532 & \textbf{0.505} \\
  \bottomrule
  \end{tabular}
\vspace{-5pt}
\end{table}


\subsection{Ablation studies}
\label{sect:ablation}
PlaneRCNN adds the following components to the Mask R-CNN~\cite{he2017mask} backbone: 1) the pixel-wise depth estimation network;
2) the anchor-based plane normal regression;
3) the warping loss module; and 4)
the segmentation refinement network.
%
To evaluate the contribution of each component, we measure performance changes while adding the components one by one.
%
Following~\cite{yang2018recovering}, we evaluate the plane segmentation quality by three clustering metrics: variation of information (VOI), Rand index (RI), and segmentation covering (SC). To further assess the geometric accuracy, we compute the average precision (AP) with IOU threshold $0.5$ and three different depth error thresholds $[0.4m, 0.6m, 0.9m]$. A larger value means higher quality for all the metrics except for VOI.

\begin{figure}[t]
	\centering
    \includegraphics[width=\linewidth]{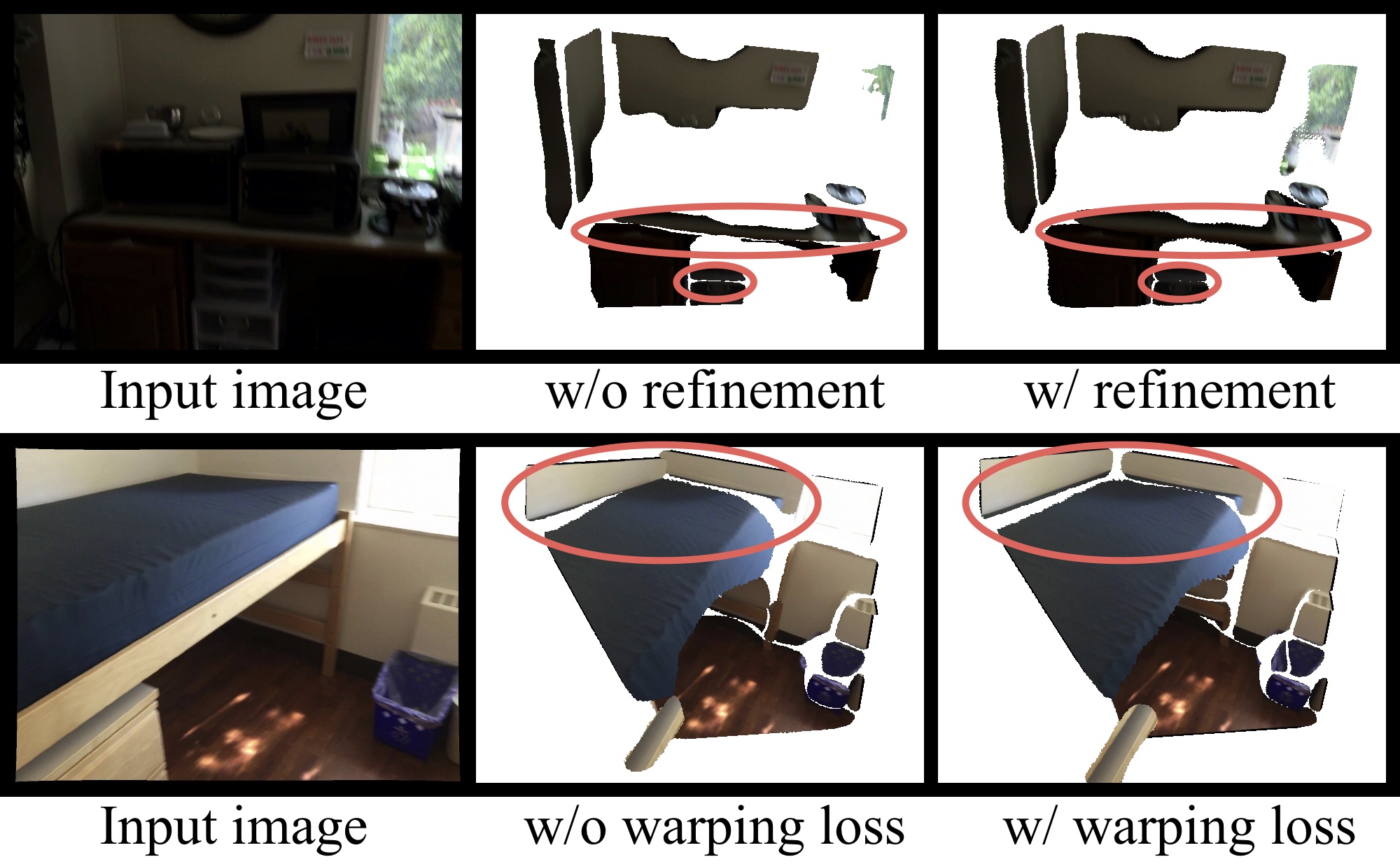}
\caption{Effects of the segmentation refinement network and the warping loss module.
Top: the refinement network narrows the gap between adjacent planes. Bottom: the warping loss helps
to correct erroneous plane geometries using the second view.}
	\label{fig:ablation}
	\vspace{3pt}
\end{figure}

\tbl{tbl:ablation} shows that all the components have a positive contribution to the final performance.
\fig{fig:ablation} further highlights the contributions of the warping loss module and the segmentation refinement network qualitatively.
The first example shows that the segmentation refinement network fills in gaps between adjacent planar regions, while
the second example shows that the warping loss module improves reconstruction accuracy with the help from the second view.

\begin{table*}[h]
\caption{Ablation studies on the contributions of the four components in PlaneRCNN. Plane segmentation and detection metrics are calculated over the ScanNet dataset. PlaneNet represents the competing state-of-the-art.
}
\vspace{2pt}
\label{tbl:ablation}
  \centering
  \begin{tabular}{l|ccc|ccc}
    \toprule
  & \multicolumn{3}{c|}{Plane segmentation metrics} & \multicolumn{3}{c}{Plane detection metrics} \\
  \midrule
Method & VOI $\downarrow$ & RI & SC & AP$^{0.4m}$ & AP$^{0.6m}$ & AP$^{0.9m}$ \\
\midrule
PlaneNet & 2.142 & 0.797 & 0.692 & 0.156 & 0.178 & 0.182 \\
\midrule
Ours (basic) & 2.113 & 0.851 & 0.719 & 0.269 & 0.329 & 0.355 \\
Ours (depth) & 2.041 & 0.856 & 0.752 & 0.352 & 0.376 & 0.386 \\
Ours (depth + anch.) & 2.021 & 0.855 & 0.761 & 0.352 & 0.378 & 0.392 \\
Ours (depth + anch. + warp.) & 1.990 & 0.855 & 0.766 & \textbf{0.365} & 0.384 & 0.401 \\
\midrule
Ours (depth + anch. + warp. + refine.) & \textbf{1.809} & \textbf{0.880} & \textbf{0.810} & \textbf{0.365} & \textbf{0.386} & \textbf{0.405} \\
  \bottomrule
  \end{tabular}
\vspace{2pt}
\end{table*}

\subsection{Occlusion reasoning}
\label{sect:occlusion}
A simple modification allows PlaneRCNN to infer occluded/invisible surfaces and reconstruct layered depthmap models. We add one more mask prediction module to PlaneRCNN to infer the complete mask for each plane instance.

The key challenge for training the network with occlusion reasoning is to generate ground-truth complete mask for supervision. In our original process, we fit planes to aligned 3D scans to obtain ground-truth 3D planar surfaces, then rasterize the planes to an image with a depth testing. We remove the depth testing and generate a ``complete mask" for each plane. Besides disabling depth checking, we further complete the mask for layout structures based on the fact that layout planes are behind other geometries. First, we collect all planes which have layout labels (e.g., \textit{wall} and \textit{floor}), and compute the convexity and concavity between two planes in 3D space. Then for each combination of these planes, we compute the corresponding complete depthmap by using the greater depth value for two convex planes and using the smaller value for two concave ones. A complete depthmap is valid if $90\%$ of the complete depthmap is behind the visible depthmap (with $0.2m$ tolerance to handle noise). We pick the valid complete depthmap which has the most support from visible regions of layout planes.



\begin{figure}[h]
	\centering
    \includegraphics[width=\linewidth]{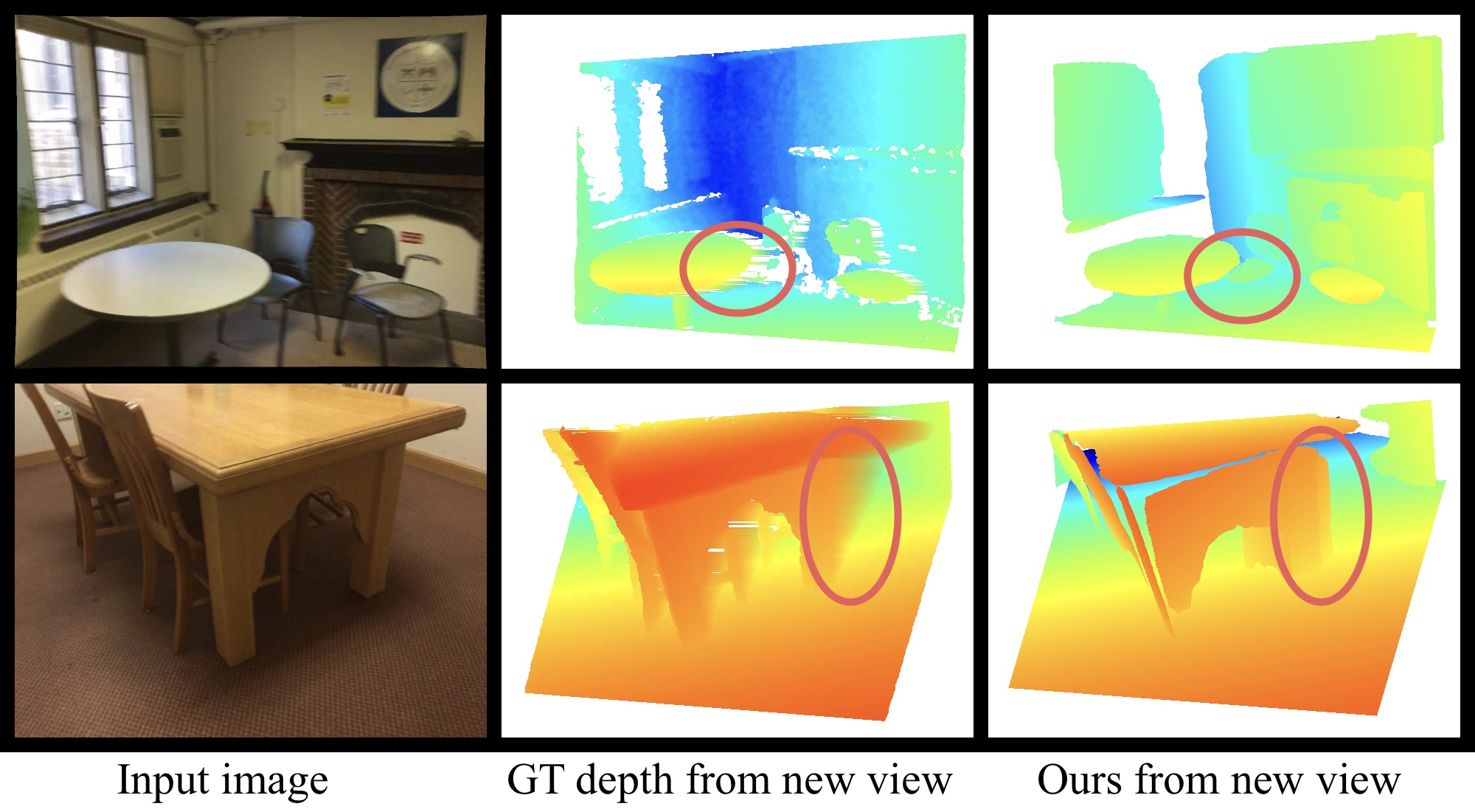}
\caption{New view synthesis results with the layered depthmap models. A simple modification allows PlaneRCNN to also infer occluded surfaces and reconstruct layered depthmap models.
}
	\label{fig:occlusion}
\end{figure}

\fig{fig:occlusion} shows the new view synthesis examples, in which the modified PlaneRCNN successfully infers occluded surfaces, for example, floor surfaces behind tables and chairs. Note that a depthmap is rendered as a depth mesh model (i.e., a collection of small triangles) in the figure.
The layered depthmap representation enables new applications such as artifacts-free view synthesis, better scene completion, and object removal~\cite{liu2016layered,tulsiani2018layer}. This experiment demonstrates yet another flexibility and potential of the proposed PlaneRCNN architecture.

\section{Conclusion and future work}
This paper proposes PlaneRCNN, the first detection-based neural network for piecewise planar reconstruction from a single RGB image. PlaneRCNN learns to detect planar regions, regress plane parameters and instance masks, globally refine segmentation masks, and  utilize a neighboring view during training for a performance boost. PlaneRCNN outperforms competing methods by a large margin based on our new benchmark with fine-grained plane annotations. An interesting future direction is to process an image sequence during inference which requires learning correspondences between plane detections.

\begin{appendices}
\section{Refinement network architecture}
\label{appendix:refinement}
\begin{figure*}
	\centering
    \includegraphics[width=0.82\linewidth]{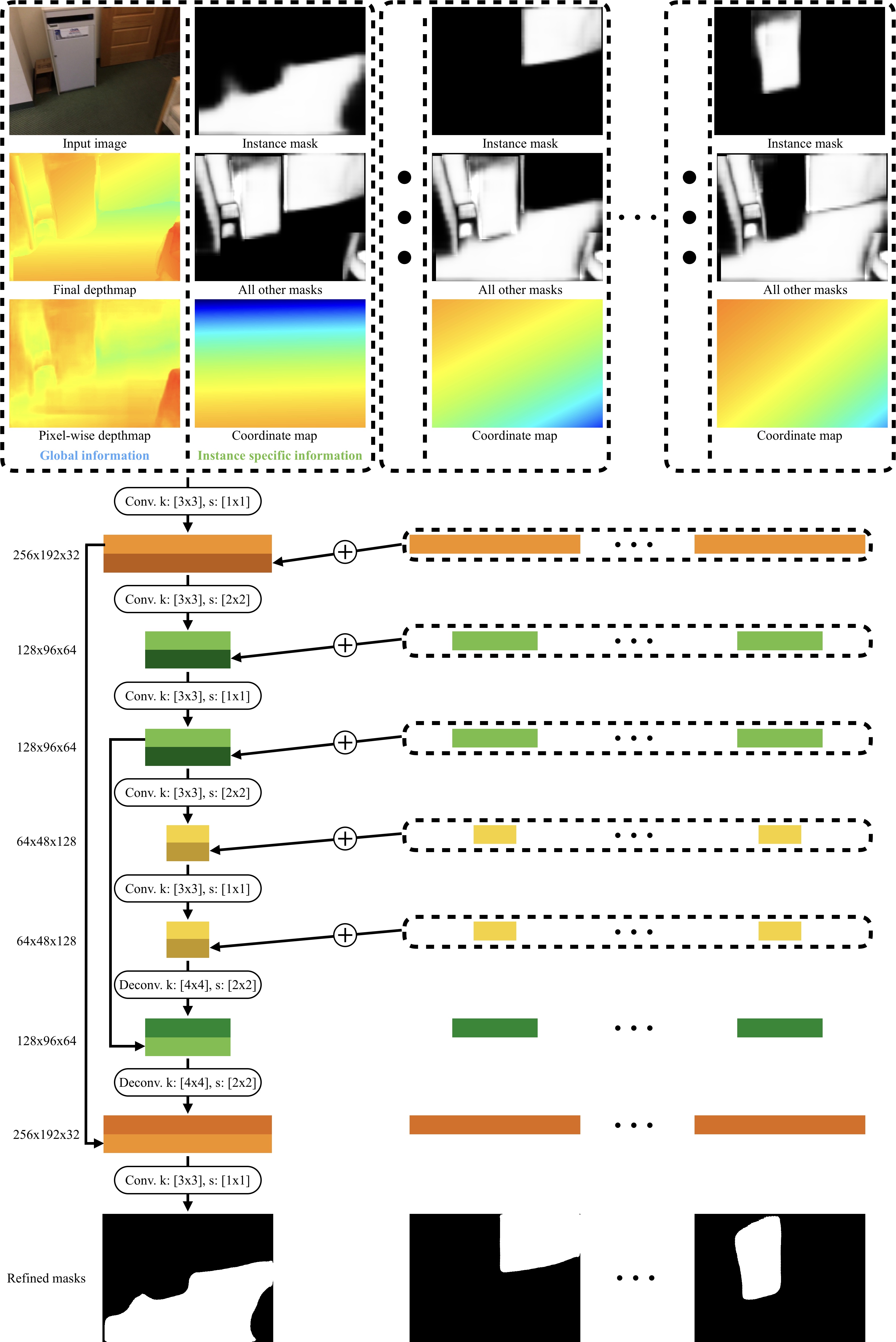}
\caption{Refinement network architecture. The network takes both global information (i.e., the input image, the reconstructed depthmap and the pixel-wise depthmap) and instance-specific information (i.e., the instance mask, the union of other masks, and the coordinate map of the instance) as input and refines instance mask with a U-Net architecture~\cite{ronneberger2015u}. Each convolution in the encoder is replaced by a ConvAccu module to accumulate features from other masks.}
	\label{fig:refinement_network}
\end{figure*}

In ~\fig{fig:refinement_network}, we illustrated the detailed architecture of the segmentation refinement network to support the description shown in ~\fig{fig:pipeline} and Sec.~\ref{sec:planerefine}.

\section{More qualitative results}
\label{appendix:results}
\begin{figure*}
	\centering
    \includegraphics[width=.95\linewidth]{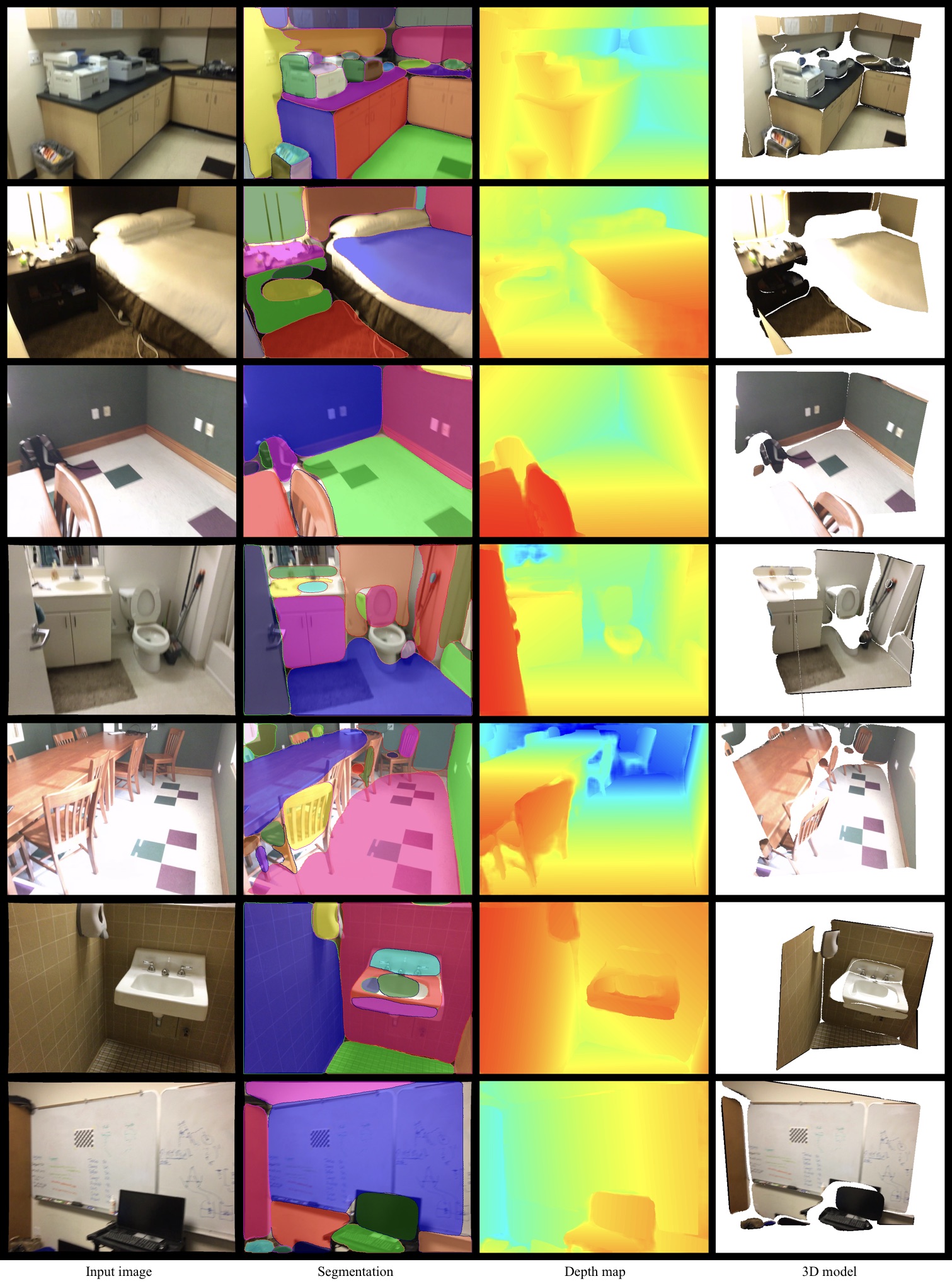}
\caption{More qualitative results on test scenes from the ScanNet dataset.}
	\label{fig:results_1}
\end{figure*}

\begin{figure*}
	\centering
    \includegraphics[width=.95\linewidth]{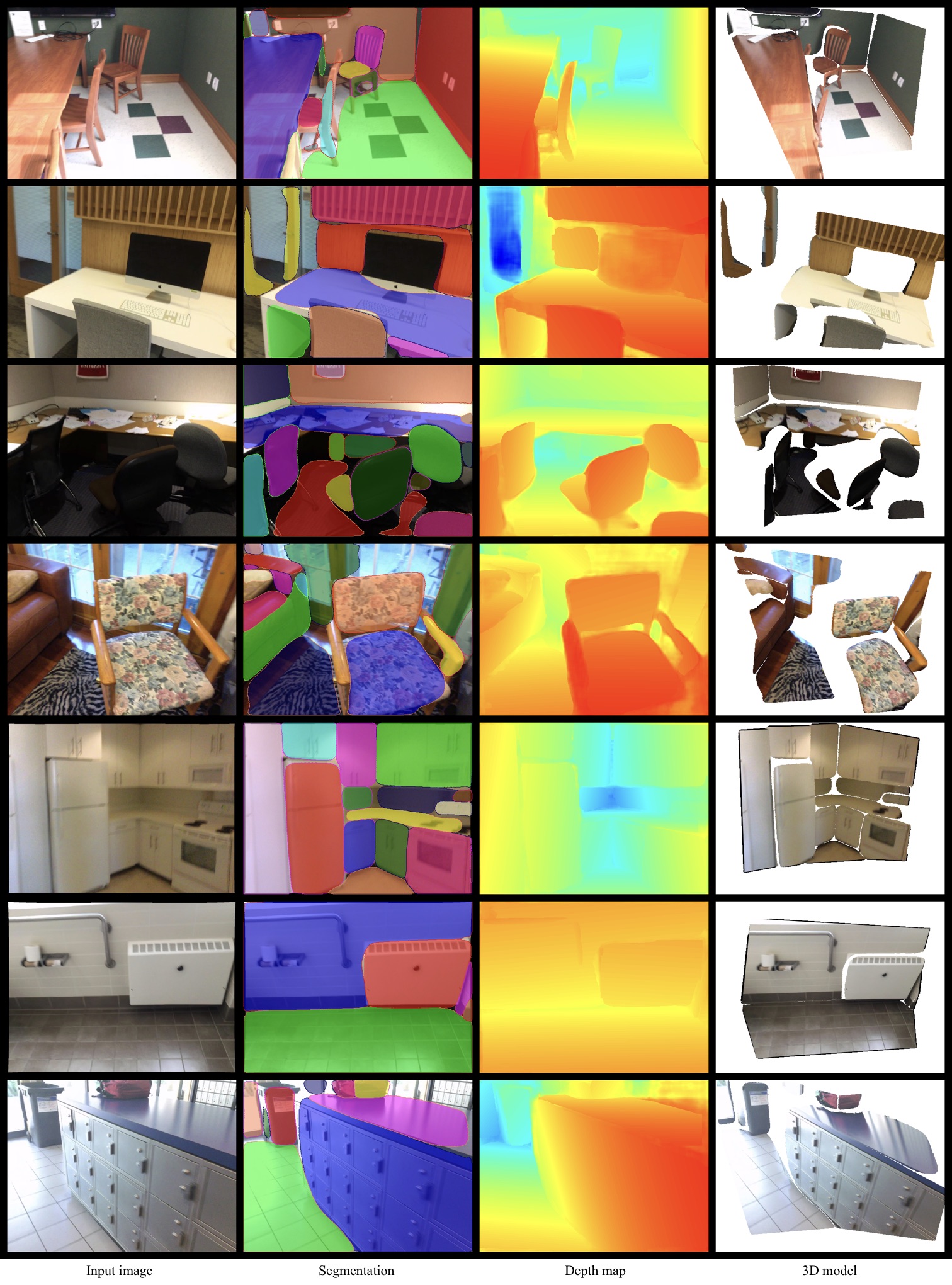}
\caption{More qualitative results on test scenes from the ScanNet dataset.}
	\label{fig:results_2}
\end{figure*}

\begin{figure*}
	\centering
    \includegraphics[width=\linewidth]{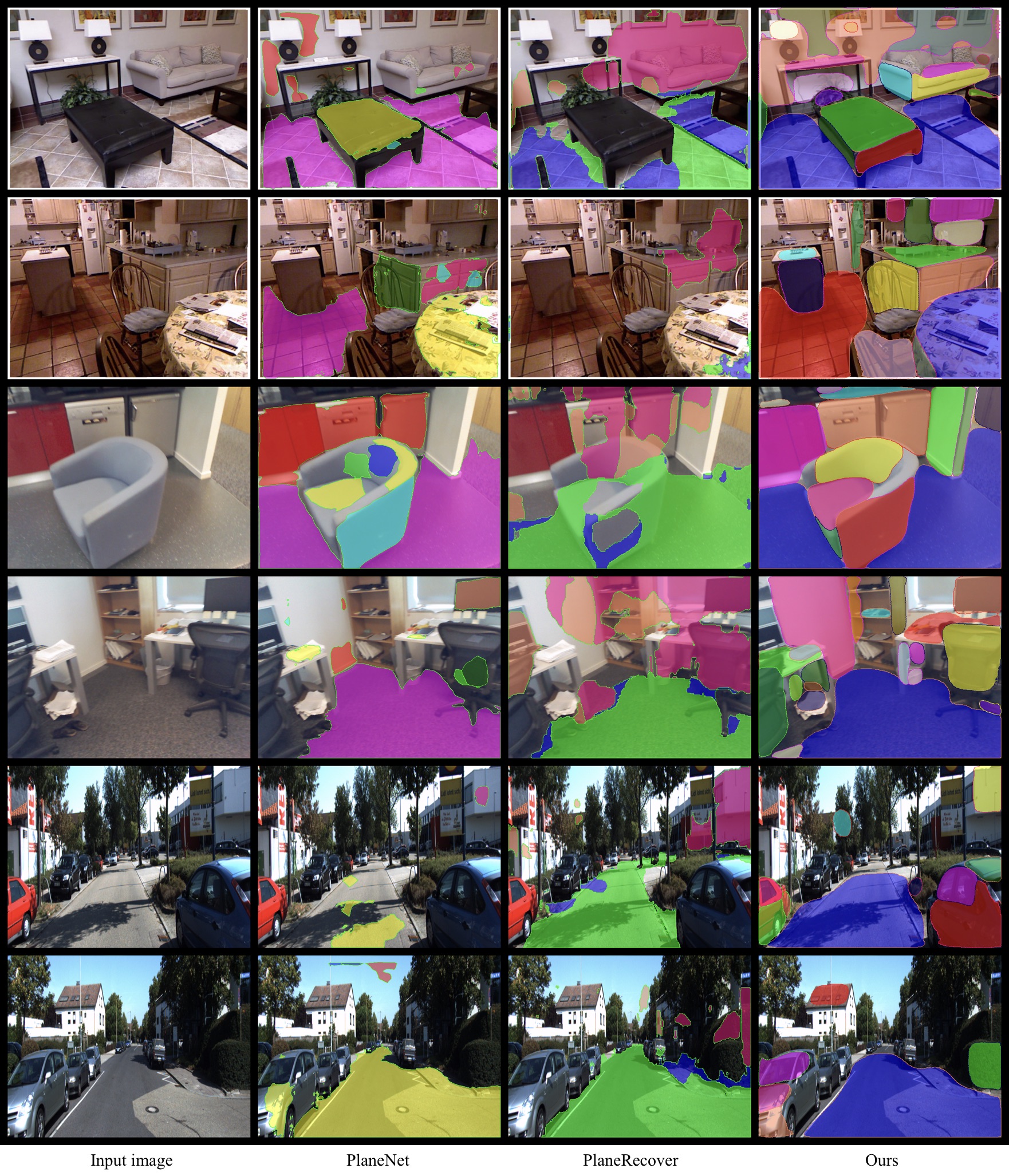}
    \caption{More plane segmentation results on unseen datasets without fine-tuning. From left to right: input image, PlaneNet~\cite{liu2018planenet} results, PlaneRecover~\cite{yang2018recovering} results, and ours. From top to the bottom, we show two examples from each dataset in the order of NYUv2~\cite{silberman2012indoor}, 7-scenes~\cite{shotton2013scene}, and KITTI~\cite{geiger2013vision}.
}
	\label{fig:cross_dataset_2}
\end{figure*}

\begin{figure*}
	\centering
    \includegraphics[width=\linewidth]{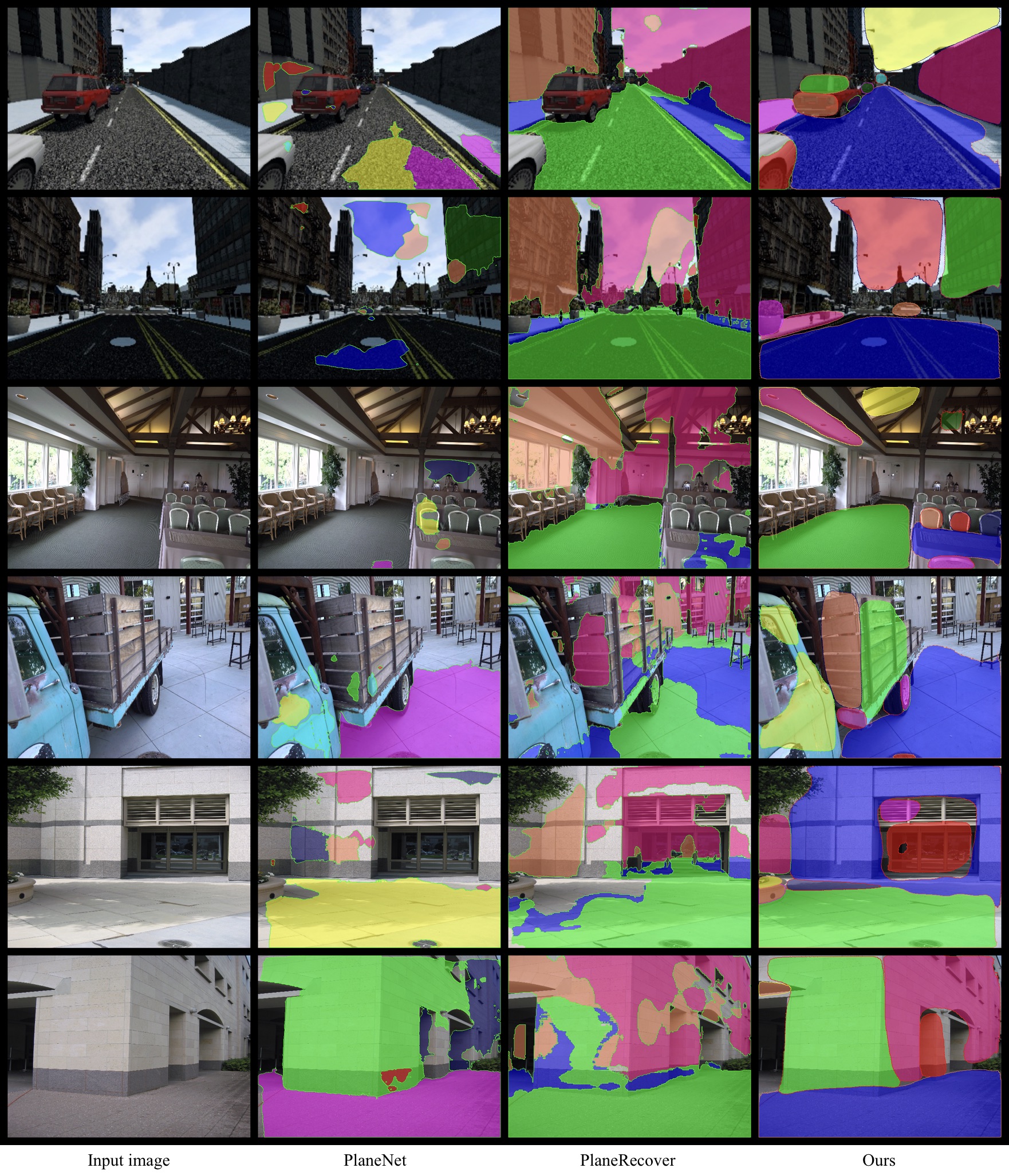}
    \caption{More plane segmentation results on unseen datasets without fine-tuning. From left to right: input image, PlaneNet~\cite{liu2018planenet} results, PlaneRecover~\cite{yang2018recovering} results, and ours. From top to the bottom, we show two examples from each dataset in the order of SYNTHIA~\cite{ros2016synthia}, Tank and Temple~\cite{knapitsch2017tanks}, and PhotoPopup~\cite{hoiem2005automatic}.
}
	\label{fig:cross_dataset_3}
\end{figure*}

We show more qualitative results of our method, PlaneRCNN, on the test scenes from ScanNet in \fig{fig:results_1} and \fig{fig:results_2}. The extra comparisons against PlaneNet~\cite{liu2018planenet} and PlaneRecover~\cite{yang2018recovering} on unseen datasets are shown in~\fig{fig:cross_dataset_2} and ~\fig{fig:cross_dataset_3}.

\end{appendices}


\end{document}